\documentclass[10pt,twocolumn,letterpaper]{article}

\usepackage{wacv}              

\usepackage{graphicx}
\usepackage{amsmath}
\usepackage{amssymb}
\usepackage{booktabs}
\usepackage{times}
\usepackage{epsfig}
\usepackage{graphicx}
\usepackage{amsmath}
\usepackage{amssymb}
\usepackage{multirow}
\usepackage{booktabs}
\usepackage{amsthm,amsfonts}

\usepackage{float}
\usepackage{multirow}
\usepackage{arydshln}
\usepackage{xcolor}
\usepackage{stfloats}
\newcommand{\myparagraph}[1]{
\vspace{0.1cm}\noindent
\textbf{#1.}
}

\usepackage[pagebackref,breaklinks,colorlinks]{hyperref}

\usepackage[capitalize]{cleveref}
\crefname{section}{Sec.}{Secs.}
\Crefname{section}{Section}{Sections}
\Crefname{table}{Table}{Tables}
\crefname{table}{Tab.}{Tabs.}

\def\wacvPaperID{1681} 
\def\confName{WACV}
\def\confYear{2024}

\begin{document}

\title{What Decreases Editing Capability? Domain-Specific Hybrid Refinement for Improved GAN Inversion}

\author{
Pu Cao$^{1,2}$\hspace{0.65cm}Lu Yang$^1$\hspace{0.65cm}Dongxv Liu$^1$\hspace{0.65cm}Xiaoya Yang$^1$\hspace{0.65cm}Tianrui Huang$^1$\hspace{0.65cm}Qing Song$^1$\thanks{corresponding author.}\\
$^1$Beijing University of Posts and Telecommunications \hspace{0.3cm}
$^2$Metavatar\\
{\tt\small \{caopu, soeaver, liudongxv, yangxiaoya, huangtianrui, priv\}@bupt.edu.cn}} 
\maketitle

\begin{abstract}
   Recently, inversion methods have been exploring the incorporation of additional high-rate information from pretrained generators (such as weights or intermediate features) to improve the refinement of inversion and editing results from embedded latent codes. While such techniques have shown reasonable improvements in reconstruction, they often lead to a decrease in editing capability, especially when dealing with complex images that contain occlusions, detailed backgrounds, and artifacts.
To address this problem, we propose a novel refinement mechanism called \textbf{Domain-Specific Hybrid Refinement} (DHR), which draws on the advantages and disadvantages of two mainstream refinement techniques. We find that the weight modulation can gain favorable editing results but is vulnerable to these complex image areas and feature modulation is efficient at reconstructing. Hence, we divide the image into two domains and process them with these two methods separately. We first propose a Domain-Specific Segmentation module to automatically segment images into in-domain and out-of-domain parts according to their invertibility and editability without additional data annotation, where our hybrid refinement process aims to maintain the editing capability for in-domain areas and improve fidelity for both of them. We achieve this through Hybrid Modulation Refinement, which respectively refines these two domains by weight modulation and feature modulation. Our proposed method is compatible with all latent code embedding methods. Extension experiments demonstrate that our approach achieves state-of-the-art in real image inversion and editing.
Code is available at \url{https://github.com/caopulan/Domain-Specific_Hybrid_Refinement_Inversion}.
\end{abstract}

\vspace{-4mm}
\section{Introduction}
\begin{figure}[h]
	\centering
	\includegraphics[width=0.95\linewidth]{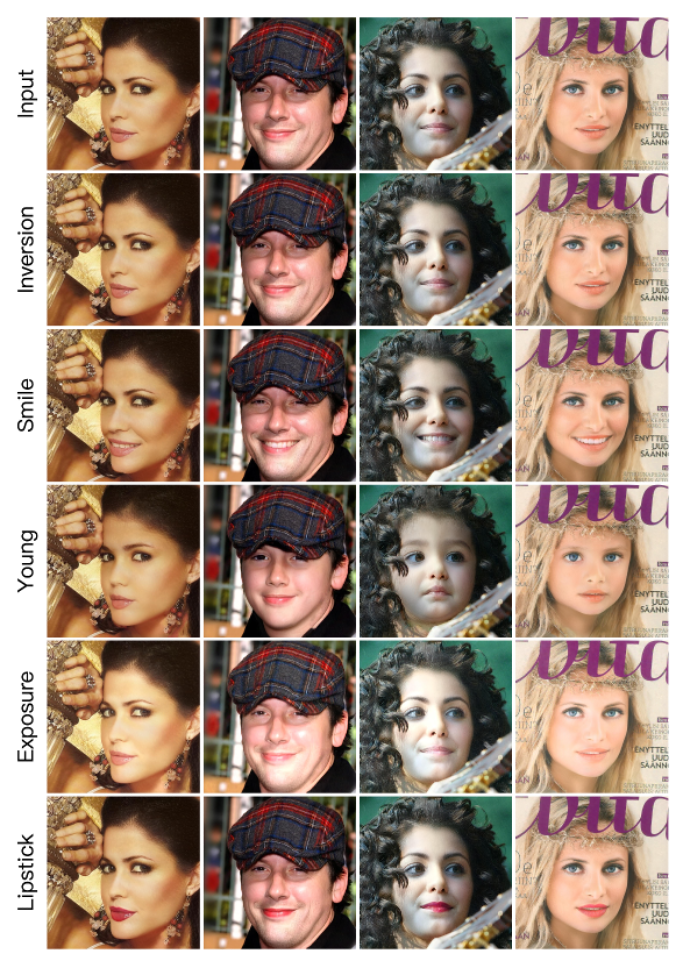}
	\vspace{-4mm}
	\caption{\textbf{Inversion and editing results of our method.} We preserve image details, including background and occlusion, in both inversion and manipulation processes.}
	\label{fig:main}
\vspace{-5mm}
\end{figure}
Generative Adversarial Networks (GANs) have shown promising results in image generation. Synthetic images are photorealistic with high resolution and are difficult to distinguish from real images \cite{karras2017progressive, karras2019style, karras2020analyzing, karras2021alias, yang2021attacks}. Based on their highly semantic latent space, image manipulation and controllable generation are deeply explored. Moreover, GANs can represent a high-quality image prior to improving various tasks, such as face parsing \cite{yang2019parsing,yang2020renovating,yang2021quality,yang2022quality,yang2022part,yang2023deep}, style transfer \cite{li2020accelerate,yang2022pastiche}, and face super-resolution \cite{wang2021towards}. However, real images are inapplicable for these applications, since most of them require latent codes in GANs' feature space.

Inversion is built to convert real images into GANs' latent space. The inverted latent codes are required to reconstruct given images by a pretrained generator, which also embeds semantic information to edit or apply in GAN-based tasks. Two types of methods generally reach image embedding. One is training an image encoder to convert given images to latent codes \cite{tov2021designing, pu2022lsap}, while another is minimizing the discrepancy between given images and reconstructed images to optimize initial latent codes iteratively \cite{karras2020analyzing}. This process attains the corresponding latent codes to reconstruct or edit the images. However, latent codes are low bit-rate \cite{wang2022high}, and high-rate details of images may not be reconstructed faithfully. Hence, many recent works focus on refining results by modulating additional high-rate information, which can be mainly divided into weight modulation and feature modulation. Weight modulation improves fidelity by iteratively tuning generator's weight by minimizing reconstruction error \cite{roich2021pivotal} or predicting weight offsets \cite{alaluf2022hyperstyle} by Hypernetworks \cite{ha2016hypernetworks}. Feature modulation encodes spatial details into intermediate features in generator, where the encoding process is also reached by iterative optimization \cite{parmar2022spatially} or model prediction \cite{wang2022high, parmar2022spatially}.

As reconstruction performance increases by refinement with high-rate information, editing capability is inevitably decreased, especially on images containing complex parts, which we show in Figure~\ref{fig:badcase}. This phenomenon is due to the destruction of the generation ability of pretrained GANs. Supervised by a discriminator, GANs are constrained to generate realistic images from sampled latent codes. However, the additional high-rate information degrades the highly semantic latent space of pretrained generator, making the latent manifold sharp by weight modulation \cite{feng2022near} or fixing the spatial distribution by intermediate features modulation to overfit the given images to reconstruct. Particulally, high-rate information needs drastic change to reconstruct complex parts. Notably, complex images prevail in the natural world. For example, face accessories, hats, occlusions, and complex backgrounds usually appear in face photos. Hence, our goal is to design a robust refinement inversion method that can faithfully reconstruct given images and retain editing capability.


Based on the above illustration, we explore the idea of ``divide and conquer"  to address this problem. Specifically, we divide the image into \emph{in-domain} and \emph{out-of-domain} parts. \emph{In-domain} parts imply areas close to generators' output distribution and are desired to perform well on both inversion and editing. Correspondingly, \emph{out-of-domain} parts are areas challenging to inverse or edit and desired to reconstruct faithfully. Hence, we introduce a hybrid refinement method to handle them. We refine  \emph{in-domain} parts by tuning generator weight since it is better to maintain editing capability. For \emph{out-of-domain} parts, we straightforwardly invert them by intermediate features to keep spatial image details. Notably, our hybrid refinement method is the first work to analyze and combine feature and weight modulation for improved GAN inversion and achieves extraordinary results as shown in Figure~\ref{fig:main}.

 Extensive experiments are presented to demonstrate the effects of our \textbf{D}omain-Specific \textbf{H}ybrid \textbf{R}efinement (DHR). We achieve state-of-the-art and gain significant improvement in both fidelity and editability. 
The key contributions of this work are summarized as follows:
\begin{itemize}
    \item We analyze the reasons for editing capability degradation in the refinement process. Based on our analysis, we introduce \emph{in-domain} and \emph{out-of-domain} and propose Domain-Specific Segmentation to segment images into these two parts for better inversion.
    \item We propose Hybrid Modulation Refinement to improve inversion results of \emph{in-domain} and \emph{out-of-domain} parts. We conduct weight modulation on \emph{in-domain} part and feature modulation on \emph{out-of-domain} part, which can preserve editing capability when refining the image details.
    \item We conduct extensive experiments and user studies to demonstrate the effects of our method. We reach extraordinary performance on real-world image inversion and editing and achieve the state-of-the-art.
\end{itemize}
\begin{figure*}[h!]
	\centering
	\includegraphics[width=1.0\linewidth]{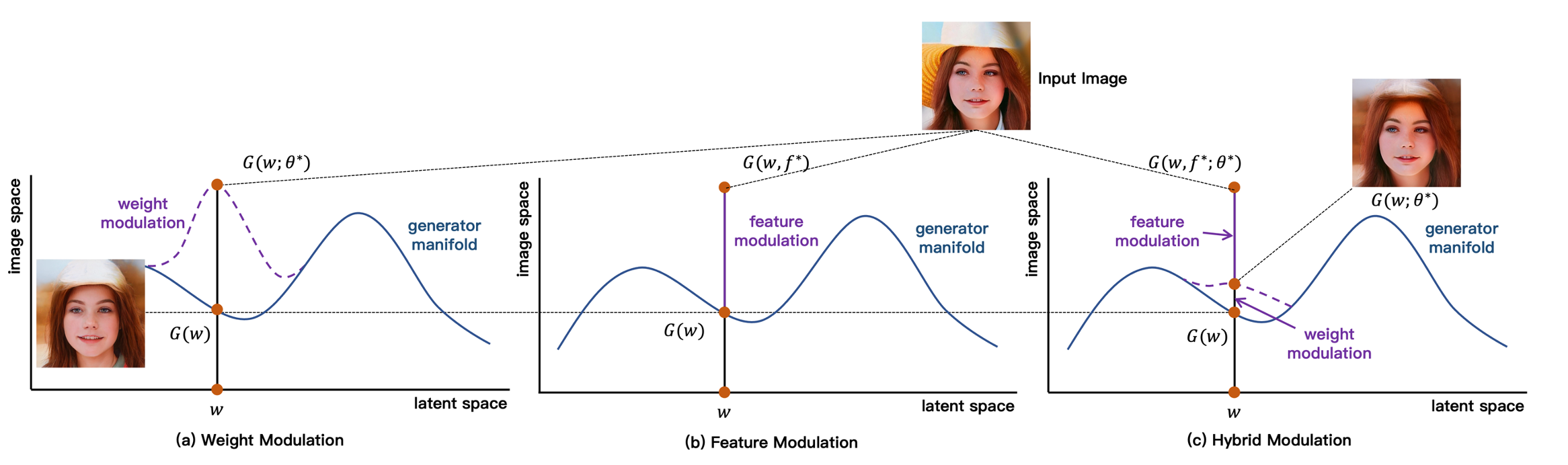}
	\caption{\textbf{Comparison of different refinement mechanisms.} We suppose that the refinement results are similar to the given images in all pipelines. The first two illustrate previous mainstream refinement mechanisms: weight, and feature modulation. Weight modulation changes the generator manifold and feature modulation introduces spatial high-rate information to recover image details. The third demonstrates our hybrid refinement method, combining these two modulation mechanisms to retain editing capability, i.e. a smooth updated manifold with image details preservation. We tune the generator on highly invertible and editable areas, which causes lower manifold deviation, and the result of this step is shown on $G(w;\theta^*)$. To reconstruct faithfully, we use feature modulation on the other area and attain $G(w,f^*;\theta^*)$.}
	\label{fig:refinement}
\vspace{-3mm}
\end{figure*}

\section{Related Work}
\subsection{GAN Inversion} 
GAN inversion aims to embed real-world images into a pretrained generator's latent space, which can be used to reconstruct and edit input images. Generally, methods can be divided into two stages. 

The first stage aims to attain low-rate latent codes, usually in $Z/W/W^+$ spaces. The latent codes are gained by an encoder or optimization process. Training an encoder \cite{tov2021designing, richardson2021encoding, wei2022e2style, guan2020collaborative, creswell2018inverting, pu2022lsap} to predict latent codes is efficient for inference and is easier to get better trade-offs between fidelity and manipulation \cite{tov2021designing,pu2022lsap}. Optimizing initial latent codes by reconstruction discrepancy gains better fidelity. However, it may cost several minutes per image \cite{karras2020analyzing, creswell2018inverting, abdal2019image2stylegan, abdal2020image2stylegan++} and decreases editability during per-image tuning. Due to low-rate property, latent codes can only reconstruct coarse information and drop the details from original images. Meanwhile, there is a trade-off between fidelity and editability, and many methods introduce additional regularization modules (e.g., latent code discriminator \cite{tov2021designing} and latent space alignment \cite{pu2022lsap}) to address it.

In the second stage, reconstruction and manipulation results from latent codes are refined by high-rate information. Refinement methods are mainly divided into weight modulation and feature modulation. Weight modulation methods predict or finetune generator weight to improve fidelity. Some methods \cite{alaluf2022hyperstyle, dinh2022hyperinverter} use hypernetwork \cite{ha2016hypernetworks} to predict weight offsets. The others tune generator by given images, which attain better fidelity but cost much time \cite{roich2021pivotal, feng2022near}. Another branch further inverts images to latent feature, which we call feature modulation. HFGI \cite{wang2022high} proposes a distortion consultation approach for high-fidelity reconstruction. SAM \cite{parmar2022spatially} segments images into various parts and inverts them into different intermediate layers by predicting ``invertibility." All of them only use one of feature and weight modulation to refine results and suffer editing capability degradation. 

\vspace{-1mm}
\subsection{GAN-based Manipulation}
GANs' latent spaces encode highly rich semantic information, which develops the GAN-based manipulation task. It aims to edit given images by changing latent codes in certain directions. Many works propose multiple methods to find semantic editing directions in GANs' latent spaces. Some methods obtain the edit vectors of the corresponding attributes by supervision with the help of attribute-labeled datasets \cite{denton2019detecting, goetschalckx2019ganalyze, spingarn2020gan}. For example, InterfaceGAN trains SVMs to find attributes' boundaries in latent space and achieve promising editing results\cite{shen2020interfacegan}. Without label annotation, other methods explore the latent space by unsupervised \cite{harkonen2020ganspace, shen2021closed, voynov2020unsupervised, wang2021geometry} or self-supervised ways \cite{jahanian2019steerability, plumerault2020controlling} to find more semantic directions way.

\section{Method}
\subsection{Preliminaries}
Inversion is built to bridge real-world images and GANs' latent space. As latent codes are low-rate, which limits their reconstruction performance, much research has recently focused on additional high-rate information in generation process, which we call refinement methods. They can be mainly divided into two categories: weight modulation and feature modulation. We first formulate them and analyze the causes of editing capacity degradation. 

\myparagraph{Formulation} We denote the original generation process as $X=G(w)$, where $G$ is the generator, and $w$ is latent code which can represent each latent space (e.g., $Z/W/W^+$). In the refinement process, we use encoded latent codes which can be attained by off-the-shelf encoders (e.g., pSp \cite{richardson2021encoding} and e4e \cite{tov2021designing}).

Weight modulation methods predict \cite{alaluf2022hyperstyle} or optimize \cite{karras2020analyzing} weight $\theta$ by minimizing reconstruction error to get $\theta^*$, and the inference process is denoted as $X=G(w;\theta^*)$. Feature modulation methods invert images into the intermediate feature, which follows $X=G(w,f)$. Defining $\mathcal{L}$ as the distance of images, we can illustrate these two refinement processes as follow:
\begin{align}
\label{equation:weight_modulation}
\mathcal{\theta^*}&=\mathop{\arg\min}_{\theta} \mathcal{L}(x,G(w;\theta)) \\
f^*&=\mathop{\arg\min}_{f} \mathcal{L}(x,G(w, f))
\end{align}

\begin{figure}[t]
	\centering
	\includegraphics[width=1\linewidth]{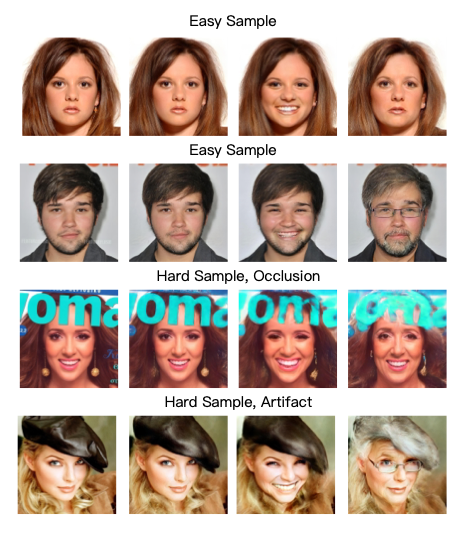}
	\vspace{-5mm}
	\caption{\textbf{Impacts on editing capability of weight modulation.} We show the input images, inversion results, and two editing results (smile and age) from PTI \cite{roich2021pivotal}. For those easy samples, editing results are reasonable. However, editing capability degrades significantly on hard samples.}
	\vspace{-4mm}
	\label{fig:badcase}
\end{figure}

\myparagraph{Impacts on editing capability} Weight and feature modulation impact image manipulation in different aspects. The schematics are shown in Figure~\ref{fig:refinement}. Since the feature modulation mechanism fixes the intermediate feature distribution at one of the layers, the effects of edit vectors applied to previous layers cannot edit the features of the latter layers. Although many existing works make efforts to maintain the editing effects, including training with adaptive distortion alignment \cite{Karras2020ada, wang2022high}, their solutions still sacrifice fidelity or editing ability \cite{parmar2022spatially}. Generally, it degrades editability more than weight modulation but is better to maintain more visual details due to high-rate spatial information.

Meanwhile, weight modulation shows promising editing performance but also gains unreasonable results on complex images, as shown in Figure~\ref{fig:badcase}. Editing results exhibit superior performance on easy samples in contrast to difficult ones. The primary cause of degradation in editing capacity is the considerable weight deviation resulting from refining complex areas, as demonstrated in Figure~\ref{fig:refinement}(a). To reconstruct given images, the weight modulation mechanism may change the generator manifold, making it much sharp on the sample point. Therefore, the highly semantic character of the pretrained generator is broken, which decreases the editing capability. 


\subsection{Domain-Specific Hybrid Refinement}
\begin{figure*}[htb]
	\centering
	\includegraphics[width=1\linewidth]{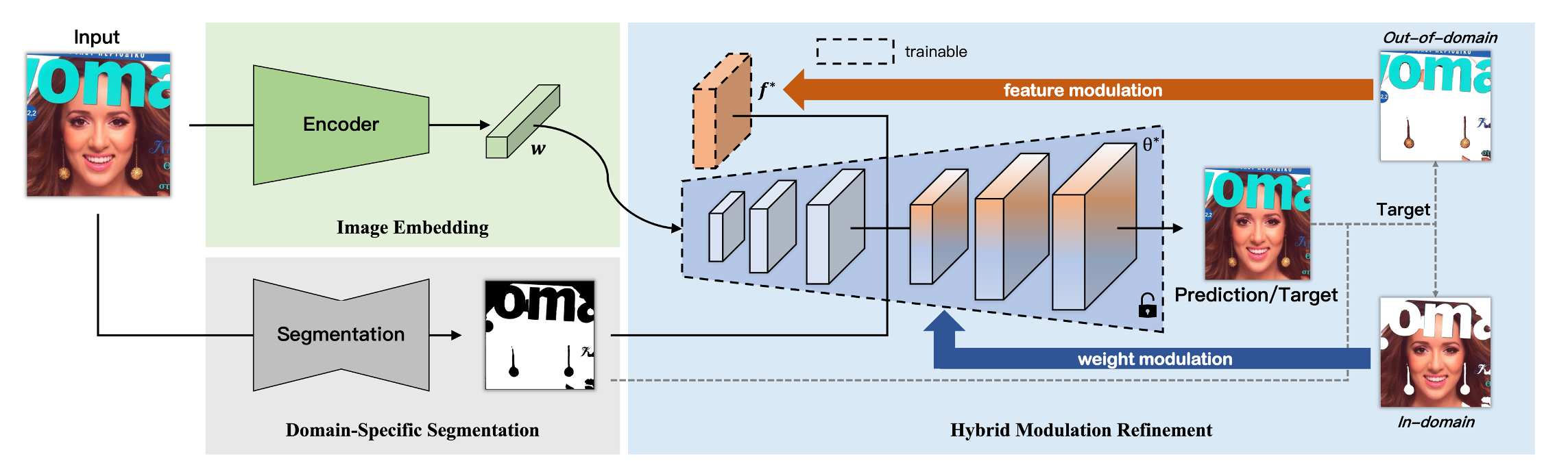}
	\caption{\textbf{Overview of our Domain-Specific Hybrid Refinement framework.} We use an off-the-shelf image encoding mechanism and introduce Domain-Specific Segmentation and Hybrid Modulation Refinement. The former segments the input images with two domains: \emph{in-domain} and \emph{out-of-domain}. They are refined by weight modulation and feature modulation in the latter method.}
	\label{fig:overview}
\end{figure*}
Based on the above analysis, weight modulation gains better editing capability but is easily affected by complex image areas, while feature modulation has better reconstruction ability. Hence, we explore the idea of ``divide and conquer" and use both of them to improve GAN inversion. In this work, we conduct \textbf{D}omain-Specific \textbf{H}ybrid \textbf{R}efinement (DHR) to deal with real-world image inversion, and the pipeline is shown in Figure~\ref{fig:overview}. 

We first propose the concepts of \emph{in-domain} and \emph{out-of-domain}. \emph{In-domain} refers to areas that have a similar distribution to the generator's output space, which makes them easy to invert. Instead, \emph{out-of-domain} areas misalign with output space, making them challenging to invert. For instance, in the face domain, \emph{in-domain} areas mainly include the face and hair, while \emph{out-of-domain} areas encompass occlusions, backgrounds, and artifacts. Additionally, \emph{in-domain} areas are more editable, such as smiling, lipstick, and eye openness.

Hence, we propose a hybrid refinement method, which segments images into \emph{in-domain} and \emph{out-of-domain} and applies weight and feature modulation to them respectively. Our framework is shown in Figure~\ref{fig:overview}, which consists of three components.

The image Embedding module aims to embed images into latent codes, which we use an off-the-shelf encoder (e.g., e4e \cite{tov2021designing} and LSAP \cite{pu2022lsap}). Given input images $X$, the encoder predicts its $W^+$ space latent codes, which we denote as $w=E(X)$, where $E$ is an encoder. 

Domain-Specific Segmentation predicts a binary mask which indicates \emph{in-domain} and \emph{out-of-domain} areas:
$$m=S(X)$$
where $m\in\{0,1\}^{H\times W}$. It segments images into two domains, which will be used for refinement.

In Hybrid Modulation Refinement, weight modulation is employed to refine \emph{in-domain} areas and restore image details for both inversion and editing results. Since the reconstruction discrepancy of \emph{in-domain} areas is low, weight deviation is limited, thereby preserving the editing capacity. Conversely, for \emph{out-of-domain} parts, we use feature modulation to refine them spatially and fix feature distribution during editing. Hence, those hard-to-invert parts do not affect editing ability. Given domain segmentation results $m$, we modulate weight $\theta$ and feature $f$ by minimizing reconstruction error in \emph{in-domain} and \emph{out-of-domain} part, respectively:
\begin{align}
\label{eq:dhr}
	\theta^*,f^*=\arg\min_{\theta,f} \mathcal{L}(X,G(w,f,m;\theta))
\end{align}

Figure~\ref{fig:refinement} illustrates the difference in the updated generator manifold between our hybrid technique and the previous refinement mechanisms. By adopting the hybrid approach, the generator manifold undergoes minimal changes and does not become too sharp at the given data point, which effectively maintains the editing capability. Notably, we conduct two refinements simultaneously and visualize $G(w,f^*;\theta^*)$ and $G(w;\theta^*)$ separately to demonstrate the effect of our method.

\subsection{Domain-Specific Segmentation}
\begin{figure}[h]
	\centering
	\includegraphics[width=1.0\linewidth]{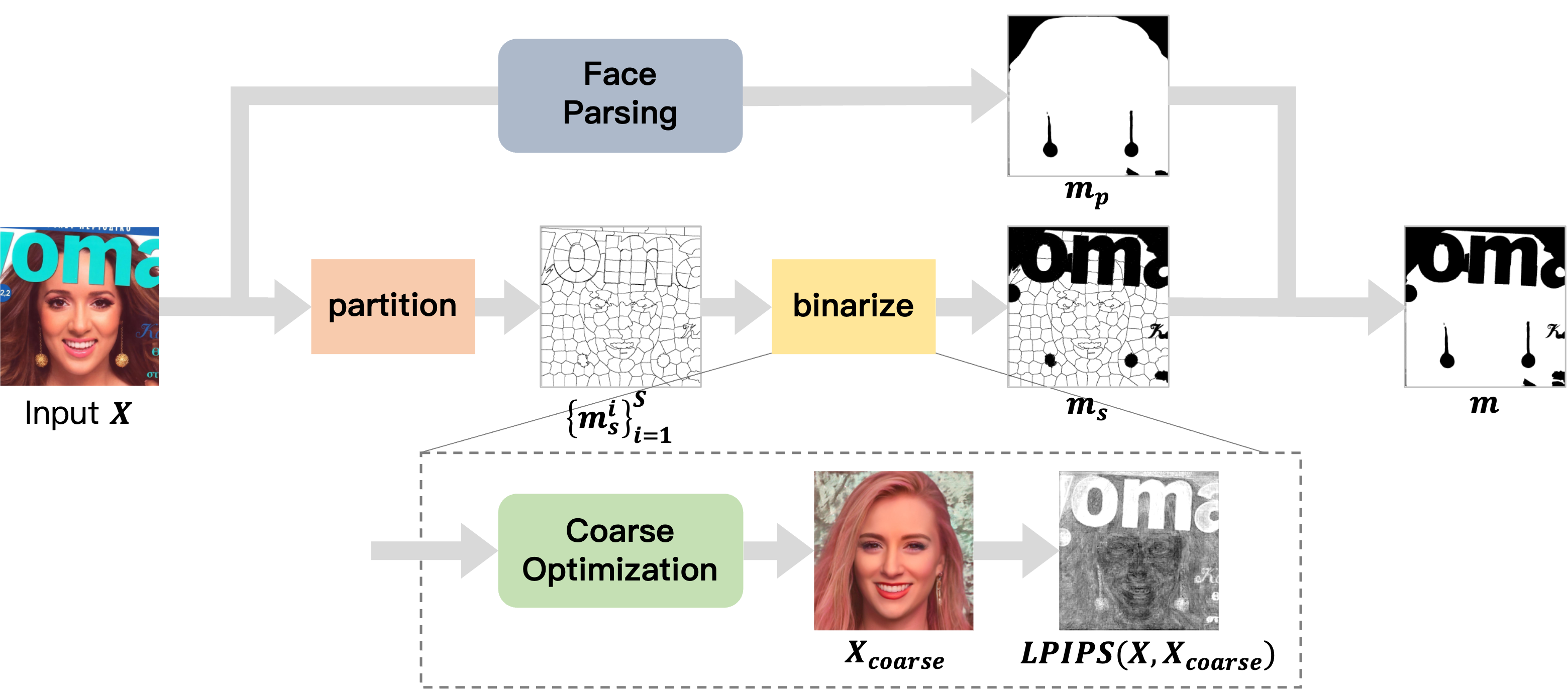}
	\caption{\textbf{Illustration of Domain-Specific Segmentation module.} We use a parsing model and superpixel algorithm with coarse optimization to segment input images into \emph{in-domain} (white areas) and \emph{out-of-domain} (black areas) parts.}
	\label{fig:dss}
	\vspace{-5mm}
\end{figure}

\begin{figure*}[h]
	\centering
	\includegraphics[width=1.0\linewidth]{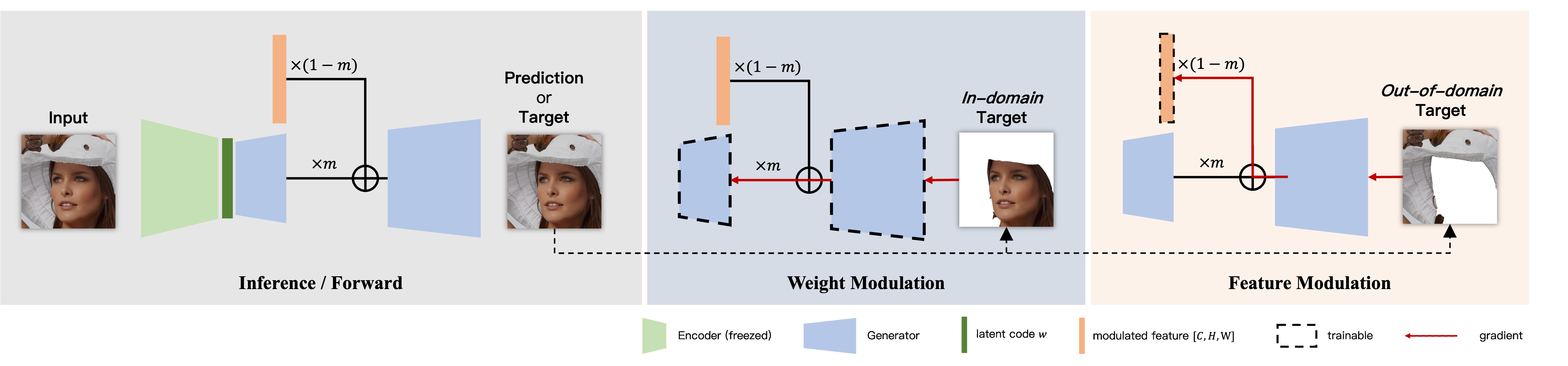}
	\caption{\textbf{Illustration of Hybrid Modulation Refinement module.} We refine \emph{in-domain} areas and \emph{out-of-domain} areas by weight and feature modulation, respectively. Black lines indicate forward flows, and \textcolor[rgb]{0.75,0,0}{red} arrows represent gradient flows.}
	\vspace{-5mm}
	\label{fig:hmr}
\end{figure*}

We propose a Domain-Specific Segmentation module to segment images into two domains: \emph{in-domain} and \emph{out-of-domain}. However, training an end-to-end learning-based domain segmentation model requires a large and annotated dataset, which can be costly. Although previous work \cite{parmar2022spatially} trains an invertibility prediction model by self-supervision, results are inaccurate in some complex areas, as we demonstrate in the appendix. To address this issue and achieve robust results on real-world images without requiring data annotation, we design an automatic segmentation pipeline consisting of two steps: ``partition" and ``binarize", which are combined with a parsing model. The pipeline is shown in Figure~\ref{fig:dss}.

We utilize a superpixel algorithm \cite{achanta2012slic} to partition the input image into multiple areas. Each partition is represented as $\{m_s^i\}_{i=1}^S$. Categorizing each partition into \emph{in-domain} and \emph{out-of-domain} without manual annotation is a crucial challenge. However, categorizing each partition into \emph{in-domain} and \emph{out-of-domain} without any manual annotation poses a significant challenge. To address this issue, we employ a coarse optimization in the $W$ space. Here, we initialize the latent codes with mean values and optimize them for a few steps. Since \emph{in-domain} are those easy-to-invert areas, the coarse inversion result $X_{coarse}$ could reconstruct \emph{in-domain} areas effectively. We compute the perceptual loss $L(X,X_{coarse})$ between the coarse reconstruction image and the input image, as shown at the bottom of Figure~\ref{fig:dss}. The white area indicates a higher loss value, while the black area represents a lower loss value. As can be observed, the loss value of the occluded area is significantly higher than that of the face area. We then calculate the average loss of each partition as follows:
$$v_i=\frac{\mathcal{L}\odot m_s^i}{||m_s^i||}$$
and binarize them by threshold $\tau$ and attain $m_{s}$.

To compensate for the missed segmentation of small areas by superpixel, we also utilize a parsing model. The parsing model categorizes face components like eye, mouth, and background \cite{zheng2021farl}. We obtain parsing results $m_p$, which we manually categorize as \emph{in-domain} or \emph{out-of-domain} based on the face components. 
Finally, domain-specific segmentation results are combined by parsing results and superpixel results:
$
m = m_p \times m_s
$.
Our Domain-Specific Segmentation module gains fine segmentation results without data annotation and is more effective for real-world images.

\subsection{Hybrid Modulation Refinement}
To maintain the editing capability of a pre-trained GAN and faithfully recover image details, we introduce a Hybrid Modulation Refinement module. This module includes two refinement aspects: weight modulation and feature modulation. Weight modulation aims to minimize \emph{in-domain} reconstruction error by tuning the generator's parameters, while feature modulation is applied to \emph{out-of-domain} areas by optimizing the intermediate feature. The forward and backward processes are shown in Figure~\ref{fig:hmr}.

For the $l$th layer of the total $k$ stages in the generator, we denote the original generator's feature as $f_l=G_{l}(w;\theta)$. An additional modulated feature is marked as $f$, which is initialized by $f_l$. When the latent codes are fixed, the original feature $f_l$ is only relevant to $\theta$. We formulate the forward process given the segmentation result $m$ as follows:
\begin{align}
f'=f_l\odot m + f\odot(1-m)
\end{align}
Then $f'$ represents the output of the first $l$ layers and generates the final images. 

For the backward process, we update weight and feature in a parallel optimization process to make them focus on the corresponding domains. We set \emph{in-domain} and \emph{out-of-domain} areas as targets for them, respectively. We use mean square error $\mathcal{L}_2$ and perceptual loss $\mathcal{L}_{lpips}$ as objectives. Calculating the reconstruction errors, we backward loss with segmentation result $m$:
\begin{align}
\mathcal{L}&=\mathcal{L}_2+\lambda\mathcal{L}_{lpips} \\
\nabla f&=\frac{\partial}{\partial f} (\frac{\mathcal{L}\odot (1-m)}{||1-m||})\\
\nabla \theta &=\frac{\partial}{\partial \theta} (\frac{\mathcal{L}\odot m}{||m||})
\end{align} 
where $\lambda$ is a hyper-parameter.
The parallel optimization mechanism constrains the impact from different domains. 
As \emph{in-domain} areas suffer small deviation, $\nabla \theta$ is lower, resulting in less effect on the generator manifold. This significantly retains its highly semantic property and preserves editing capability.


\begin{figure*}[h]
	\centering
	\includegraphics[width=1.0\linewidth]{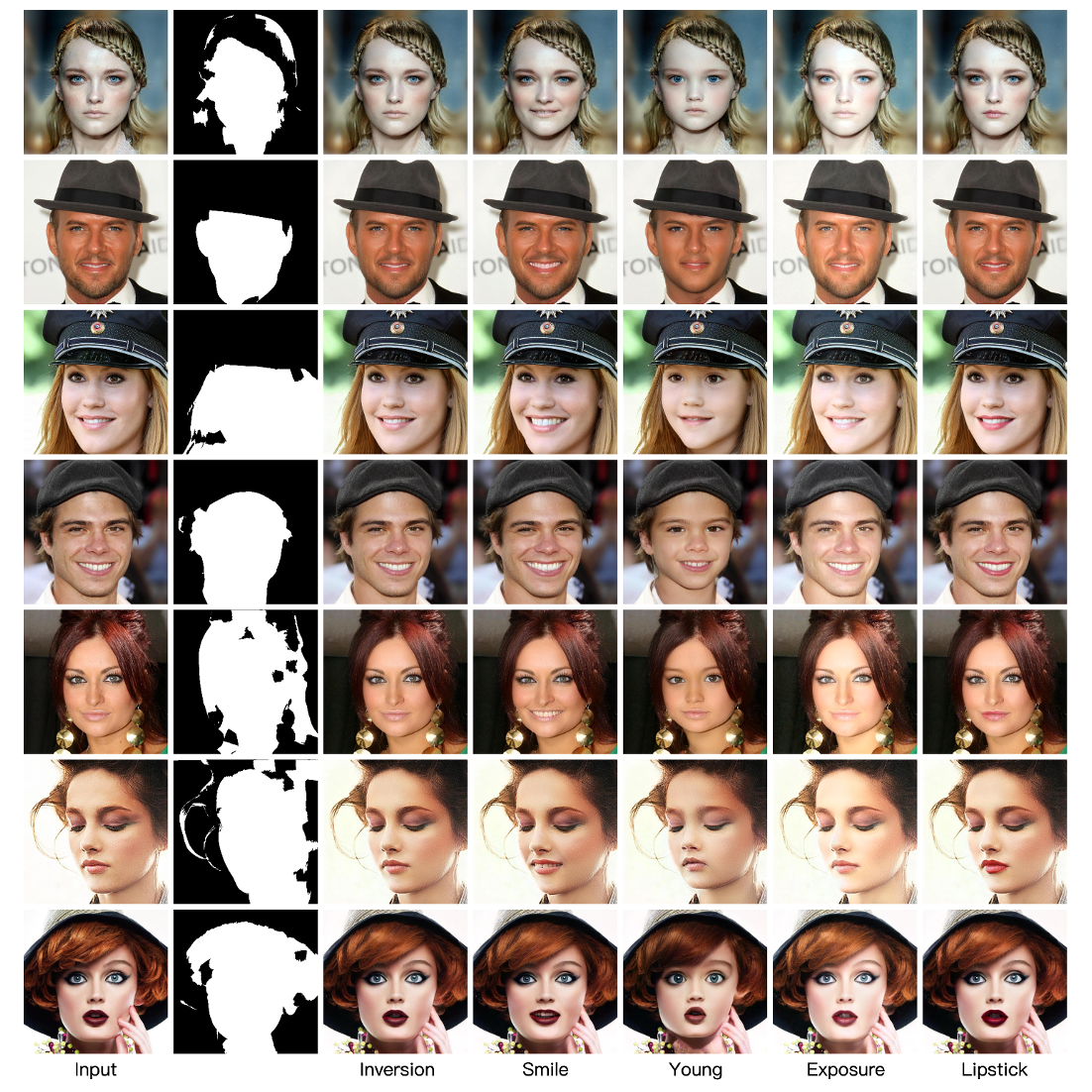}
	\caption{\textbf{Inversion and editing results.} The second column is the segmentation results of \emph{in-domain} and \emph{out-of-domain} areas. Our method restores almost all image details.}
	\label{fig:result}
\vspace{-5mm}
\end{figure*}

\section{Experiments}
\subsection{Experimental Settings}

\myparagraph{Datasets} We evaluate all methods on the CelebA-HQ \cite{karras2017progressive,liu2015deep} test set (2,824 images). Encoders and the generator are trained on FFHQ \cite{karras2019style} (70,000 images).

\myparagraph{Baselines} We compare our model to previous state-of-the-art refinement methods, i.e., ReStyle \cite{alaluf2021restyle}, HFGI \cite{wang2022high}, SAM \cite{parmar2022spatially}, and PTI \cite{roich2021pivotal}. We use pSp \cite{richardson2021encoding}, e4e \cite{tov2021designing} and LSAP \cite{pu2022lsap} as encoders. Moreover, the performance of encoder-based methods is also reported. All of weights of encoders come from their official release.

\myparagraph{Metrics} We evaluate all methods at two respects: inversion and editing. For inversion ability, we conduct MSE, LPIPS \cite{zhang2018unreasonable}, and identity similarity calculated by a face recognition model \cite{huang2020curricularface}. MSE straightforwardly measures the image distortion, and LPIPS evaluates the visual discrepancy. Identity similarity further compares the identity consistency during inversion. Moreover, we perform user studies to evaluate the perceptual performance of inversion and editing. 
 
\subsection{Main Results}
\myparagraph{Quantitative results} We first evaluate the reconstruction ability quantitatively. The results are reported in Table~\ref{table:face_inversion}. We compare our method with four previous refinement methods and employ two encoders. For PTI, we conduct experiments with encoders and their proposed $W_{pivot}$. As one can see, DHR achieves the best performance on all metrics. Employed with e4e, it gains $0.0036$ MSE, which is $7.5\%$ of the vanilla e4e, $8.3\%$ of ReStyle, $17\%$ of HFGI, $25\%$ of SAM, and $48\%$ of PTI. For LPIPS and identity similarity, it demonstrates similar superiority and surpasses other methods by a large margin. 

This is because our hybrid refinement approach attains an extraordinary trade-off between fidelity and editability. The previous feature modulation methods (i.e., HFGI and SAM) constrain the effects of the additional feature information by elaborate fusion module or feature regularization to preserve editing capability, and Restyle and PTI encode spatial details in low-rate latent codes or global convolutional weights (hard to recover spatial visual details). All of them employ stronger limitations on high-rate information to maintain editing capability. We will then demonstrate that our DHR not only gains better quantitative reconstruction results but also attain better editing results without any regularization.

\myparagraph{Qualitative results} We illustrate the inversion and editing results of DHR in Figure~\ref{fig:result}. The second column is the results from the Domain-Specific Segmentation module, and the third is our inversion results. Some dedicated areas are categorized into \emph{out-of-domain} domain (black area), such as the braid area in the first image. We edit them in four directions, i.e., smile, young, exposure, and lipstick. All image details are preserved in both inversion and editing results, such as hairstyle (the first row), earrings (the fifth row), and hats. 

\begin{figure*}[t!]
	\centering
	\includegraphics[width=0.95\linewidth]{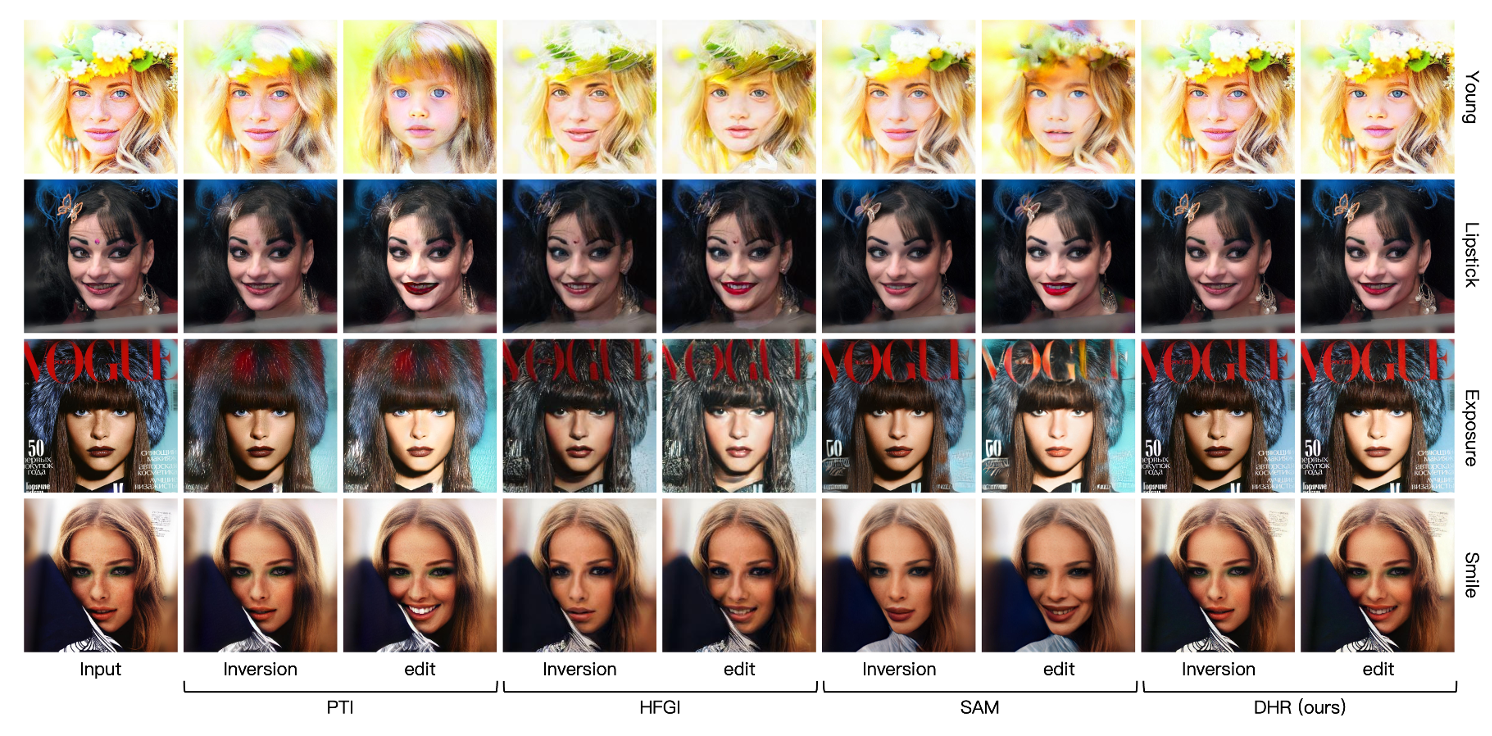}
	\vspace{-3mm}
	\caption{\textbf{Comparisons to previous methods.} We compare the inversion and editing results with PTI \cite{alaluf2021restyle}, HFGI \cite{wang2022high}, and SAM \cite{parmar2022spatially}. Although HFGI and SAM reach reasonable inversion results, image distortion and details loss occur in editing results. Our method attains the best fidelity and editing performance. Image details are reserved in both phases, and our results are the most natural.}
	\label{fig:comparison}
\vspace{-3mm}
\end{figure*}

We further conduct a qualitative comparison with other methods, shown in Figure~\ref{fig:comparison}. Although inversion results are reasonable, editing capability degradation occurs in baselines, e.g., decorations and occlusion blur. Our editing results are more faithful and photorealistic.

\begin{table}[t]
\resizebox{0.45\textwidth}{!}
{

\begin{tabular}{llccc}
\toprule
\textbf{Method}                    &   \textbf{Encoder}  & \textbf{MSE} $\downarrow$                 & \textbf{LPIPS} $\downarrow$               & \textbf{Similarity} $\uparrow$            \\ \midrule
\multirow{2}{*}{ReStyle \cite{alaluf2021restyle}} & pSp & 0.0276                           & 0.1298                           & 0.5816                           \\
      & e4e & 0.0429                           & 0.1904                           & 0.5062 \\ \cdashline{1-5}
\multirow{2}{*}{HFGI \cite{wang2022high}} & e4e & 0.0210                           & 0.1172                           & 0.6816                           \\
      & LSAP & 0.0210                           & 0.0945                           & 0.7405 \\ \cdashline{1-5}
\multirow{2}{*}{SAM \cite{parmar2022spatially}}               & e4e & 0.0143                           & 0.1104                           & 0.5568                           \\
 & LSAP & 0.0117 & 0.0939 & 0.6184                           \\\cdashline{1-5}
\multirow{3}{*}{PTI \cite{roich2021pivotal}} & e4e & 0.0074 & 0.0750 & 0.8633  \\
 & LSAP  & 0.0067 & 0.0666 & 0.8696   \\ 
 & $W_{pivot}$ & 0.0084 & 0.0845 & 0.8402 \\
\cdashline{1-5}
 \multirow{2}{*}{DHR (ours)} & e4e & 0.0036 & 0.0455 & 0.8704                           \\
 & LSAP & \textbf{0.0035} & \textbf{0.0436} & \textbf{0.8780}   \\ \hline
  \multirow{3}{*}{\textcolor{gray}{Encoder-Only}}                      & \textcolor{gray}{pSp} \cite{richardson2021encoding}   & \textcolor{gray}{0.0351} & \textcolor{gray}{0.1628} & \textcolor{gray}{0.5591} \\
                      & \textcolor{gray}{e4e} \cite{tov2021designing}   & \textcolor{gray}{0.0475}                          & \textcolor{gray}{0.1991}                           & \textcolor{gray}{0.4966}                           \\
                  & \textcolor{gray}{LSAP} \cite{pu2022lsap}   & \textcolor{gray}{0.0397}                           & \textcolor{gray}{0.1766}                           & \textcolor{gray}{0.5305} \\
\bottomrule
\end{tabular}
}
\vspace{-2mm}
\caption{\textbf{Fidelity results on face domain.} We compare DHR to three previous refinement methods with two powerful encoders. The results of these encoders are also presented at the bottom.}
\vspace{-2mm}
\label{table:face_inversion}
\end{table}

\subsection{User Study}
\begin{table}[]
\resizebox{0.45\textwidth}{!}
{
\begin{tabular}{lccccc}
\toprule
\multirow{2}{*}{Method} & \multirow{2}{*}{Inversion} & \multicolumn{4}{c}{Editing}         \\ \cline{3-6} 
       &           & Smile & Young & Exposure & Lipstick \\ \hline
ReStyle \cite{alaluf2021restyle}    &     94\%      &    100\%   &   100\%    &      90\%    &      94\%    \\

HFGI \cite{wang2022high}  &    94\%     &    84\%   &    86\%   &     100\%     &          96\% \\
SAM \cite{parmar2022spatially}   &     100\%      &   92\%       &   94\%    &    100\%      &     100\%     \\ 
PTI \cite{roich2021pivotal}    &     96\%      &    100\%   &   100\%    &      84\%    &      88\%    \\
\bottomrule
\end{tabular}
}
\vspace{-3mm}
\caption{\textbf{User study.} We conduct user studies on inversion and editing tasks. The values in the table indicate the percentage of images where users prefer our results. Results show our method is more faithful and photorealistic.}
\label{tab:user_study}
\vspace{-3mm}
\end{table}

We conduct user studies to demonstrate the performance of inversion and editing. Results are shown in Table~\ref{tab:user_study}. We randomly select 50 different images and invert and edit them by HFGI \cite{wang2022high}, SAM \cite{parmar2022spatially}, PTI \cite{roich2021pivotal}, and our method. We then ask three users to make a preference for each pair of images. A higher value implies users prefer our results. As can be seen, our results are highly preferred by users, which are most all above 90\%, compared to previous state-of-the-art methods. It illustrates that our method decreases image distortion and attains better photorealism of reconstruction and manipulation results.

\section{Conclusion}
In this work, we present a novel inversion approach, Domain-Specific Hybrid Refinement, aimed at improving GAN inversion and editing capability. We investigate the causes of editing ability degradation in the refinement process and introduce a ``divide and conquer" strategy to address this issue. Our method consists of two main components: Domain-Specific Segmentation and Hybrid Modulation Refinement. The former segments images into \emph{in-domain} and \emph{out-of-domain} parts without data annotation, while the latter refines them by weight and feature modulation, respectively. Our method achieves promising results in both inversion and editing tasks, with significant improvements over existing approaches.

\small\noindent\textbf{Acknowledgements}
This work was supported by the National Key Research and Development Program of China (Grant No. 2022YFC3302200), China Postdoctoral Science Foundation (2022M710467), National Key Research and Development Program of China (Grant No. 2021YFF0500900), and Intelligent Logistics Interdisciplinary Team Project of BUPT.

{\small
\bibliographystyle{ieee_fullname}
\bibliography{egbib}
}

\def\maketitlesupplementary
   {
   \newpage
       \twocolumn[
        \centering
        \Large
        \textbf{What Decreases Editing Capability? Domain-Specific Hybrid Refinement for Improved GAN Inversion}\\
        \vspace{0.5em}Supplementary Material \\
        \vspace{1.0em}
       ] 
   }
   
\clearpage
\maketitlesupplementary
\appendix
\section{Implementation Details}
\myparagraph{Architecture} In the Domain-Specific Segmentation module, we use FaRL \cite{zheng2021farl} and SLIC \cite{achanta2012slic} as the parsing model and superpixel algorithm. We use $1024\times1024$ StyleGAN2 as face generator, which is trained on FFHQ  dataset.

\myparagraph{Hyper-parameters} In Domain-Specific Segmentation module, we set two thresholds $\tau_1=0.7$ and $\tau_2=0.8$, according to parsing results, respectively. We set a higher threshold for eyes, nose, and mouse areas. For In Hybrid Modulation Refinement, coarse inversion optimizes initial latent codes for 40 steps. We add the additional modulation feature at $11$th layer amount 18 layers of $1024\times1024$ StyleGAN2. Adam optimizers with no weight decay and betas$(0.9, 0.999)$ are used for weight and feature modulation, and learning rates are $0.0015$ and $0.09$, respectively. The feature is optimized in 100 steps and weight is only optimized in 50 steps for better editing results. 


\section{Runtime Analysis}
\begin{figure}[h]
	\centering
	\includegraphics[width=1.0\linewidth]{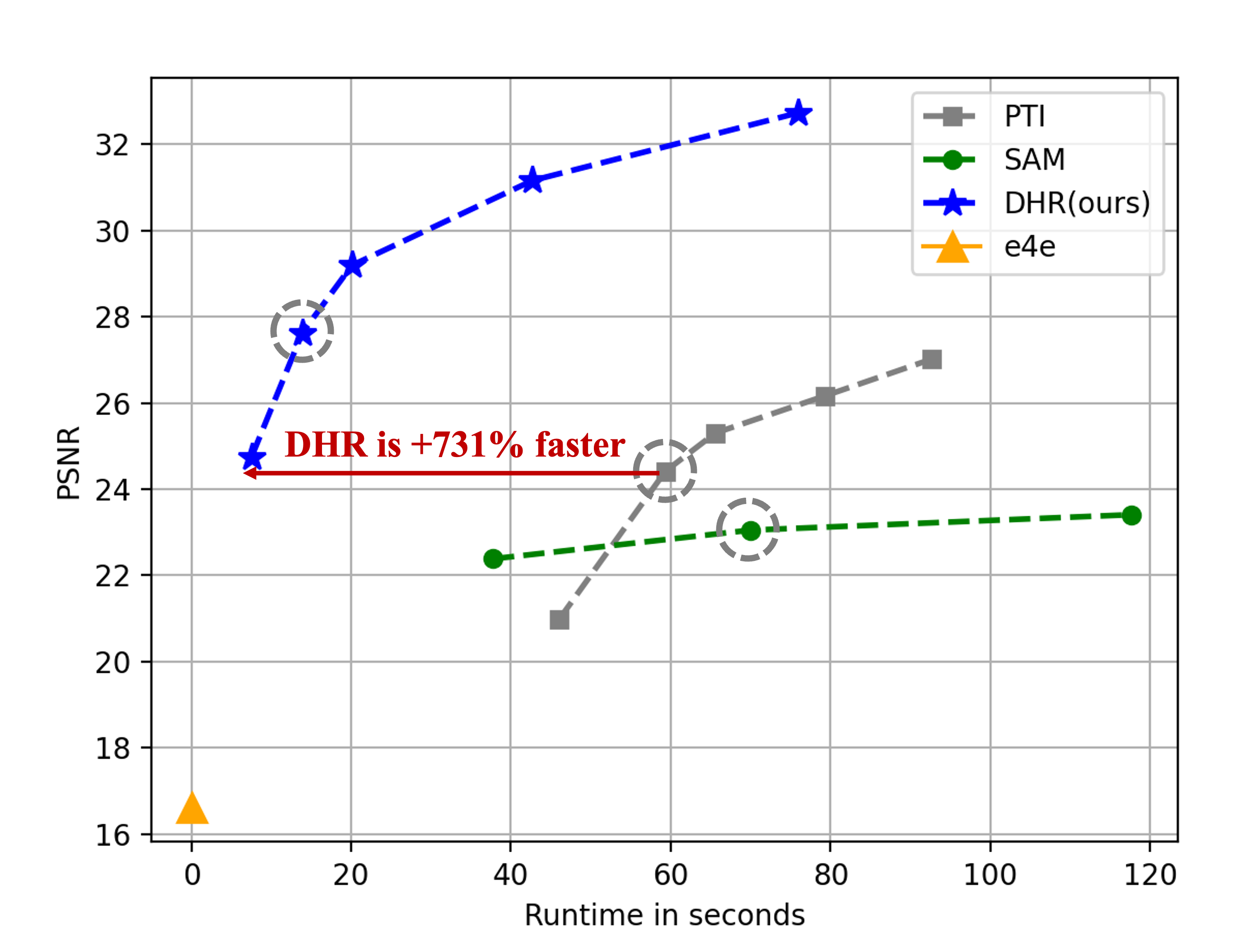}
	\caption{\textbf{Runtime Analysis.} We compare three optimization-based refinement methods using 100 CelebA-HQ test images. The default configuration of each method is marked by a dashed box. Our method greatly accelerates the refinement optimization process and gains much higher performance. \textbf{DHR only cost 7.5s to achieve the previous SOTA.}}
	\label{fig:runtime}
\end{figure}
One main concern of optimization-based methods is time-consuming. Previous methods (i.e., PTI and SAM) cost more than one minute per image, which is hardly applied. We next demonstrate that DHR dramatically outperforms all existing methods in terms of speed. 

\myparagraph{Settings} As PTI \cite{roich2021pivotal}, SAM \cite{parmar2022spatially} and DHR are optimization-based refinement inversion methods and gain better inversion results, we compare their runtime and PSNR using 100 randomly selected images from CelebA-HQ test set. All experiments are conducted on a single NVIDIA RTX 3090 GPU and i9-10900X CPU (superpixel algorithm using CPU). For PTI and SAM, we change the optimization steps to get each data point, while the optimization process for pivotal latent codes in PTI is fixed. For DHR, we fix the coarse inversion steps (i.e., 40 steps), and change the feature modulation steps while weight modulation steps are set to half. The result is shown in Figure~\ref{fig:runtime}. 

\myparagraph{Results} Compared to previous methods, DHR accelerates the refinement process significantly and attains higher performance (almost 33dB of PSNR). Although the overly long modulation process may cause editing capability degradation, we experimentally find that 200/100 steps (20.1s with 29.19dB PSNR) for feature/weight modulation gain reasonable results. \textbf{Our method achieves previous SOTA using only 7.5s, compared to 62.3s for PTI, and gains extroadinary results in 14s}.

\section{Effects of Domain-Specific Segmentation}
\begin{figure*}[h]
	\centering
	\includegraphics[width=1.0\linewidth]{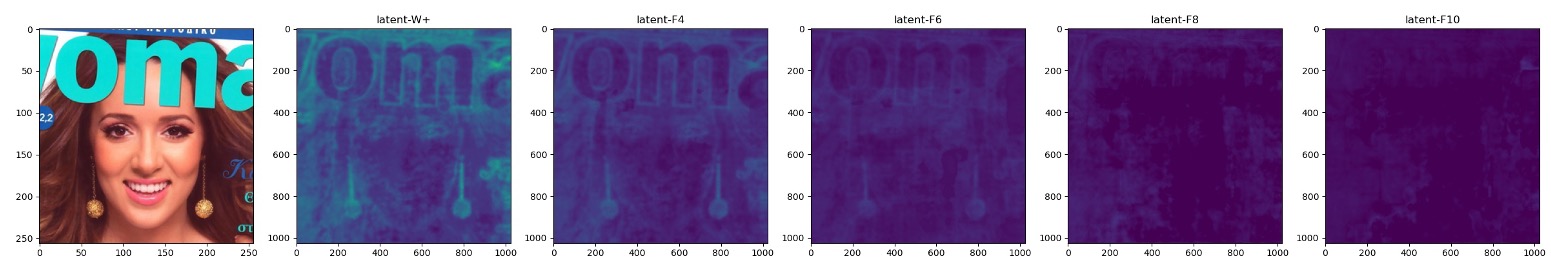}
	\caption{\textbf{Invertibility prediction in SAM\cite{parmar2022spatially}.}}
	\label{fig:invertibility}
\end{figure*}

\begin{figure}[h]
	\centering
	\includegraphics[width=1.0\linewidth]{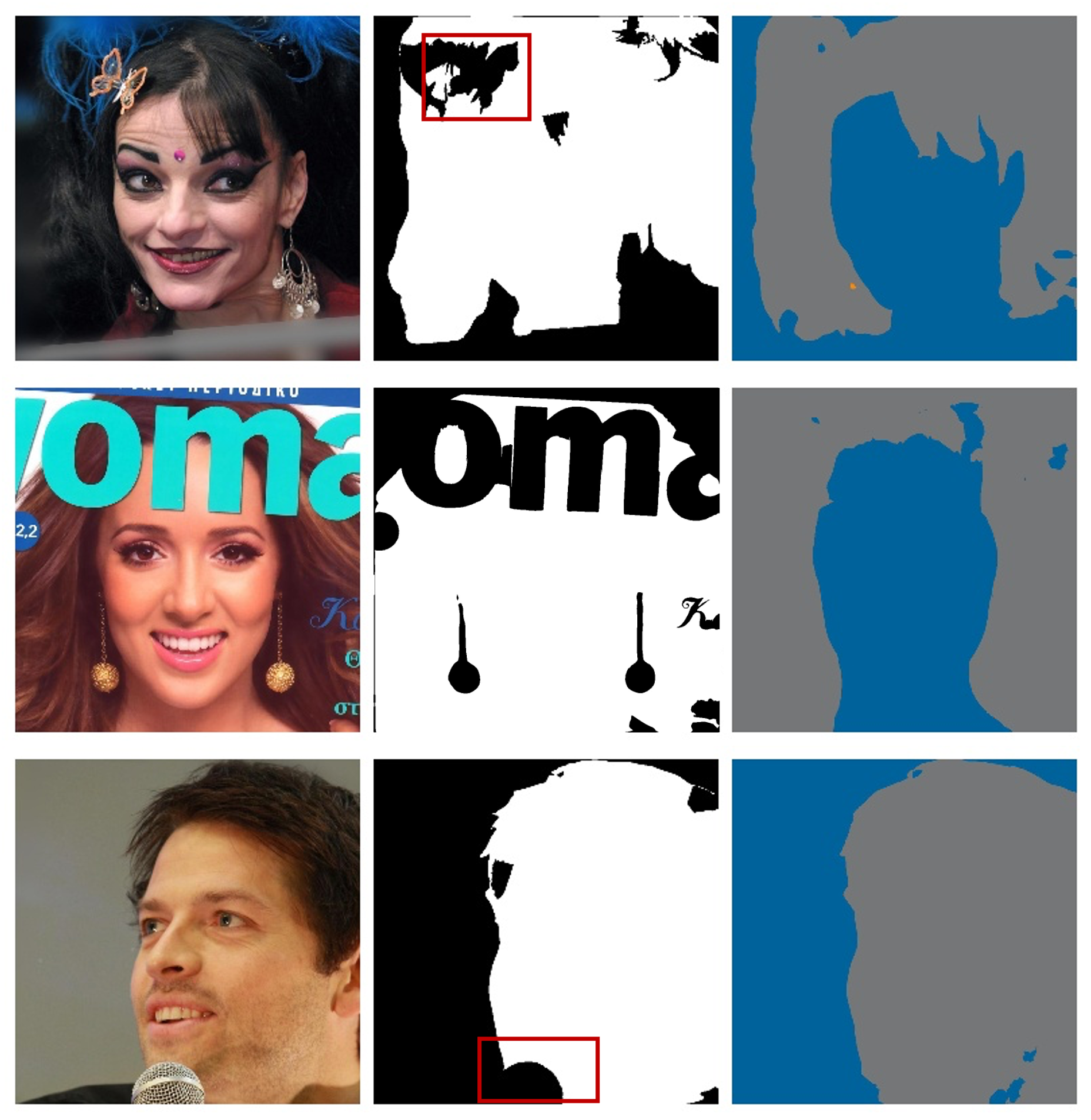}
	\caption{\textbf{Comparison between Domain-Specific Segmentation and Invertibility Segmenter in SAM\cite{parmar2022spatially}.}}
	\label{fig:ablation}
\end{figure}

We illustrate the difference between Domain-Specific Segmentation (DSS) and invertibility segmenter in SAM\cite{parmar2022spatially} in Figure~\ref{fig:ablation}. As invertibility segmenter consists of invertibility predictor and pretrained segmentation model, it has two primary differences from DSS: 

\myparagraph{Semantic prior plays a significant role in invertibility prediction} Since there is no semantic information, simple color block areas are easily misclassified, as can be seen in Figure~\ref{fig:invertibility}. The characters  have high invertibility in $\mathcal{W}^+$ space. Our DSS draws on the semantic  face prior from pretrained GAN, having better ability to distinguish ``easy-to-invert" (i.e., \emph{in-domain}) and ``hard-to-invert" (i.e., \emph{out-of-domain}).

\myparagraph{General segmentation model can not segment complex area} Another main difference is that SAM uses general segmentation models to split areas, which can only distinguish pre-defined categories. In real-world images, there are a large number of complex areas, such as hairpins, occlusion, and microphones. We address it by using the superpixel algorithm, which is required to partition image into different areas and is category-agnostic.

This also verifies the idea of ``partition and binarize". We use superpixel to split tens of areas and coarsely inverse to binarize each of them. Hence, DDS is robust to real-world images and attains promising segmentation results.

%
%
%

\section{Additional Results}
\begin{figure*}[t]
	\centering
	\includegraphics[width=1.0\linewidth]{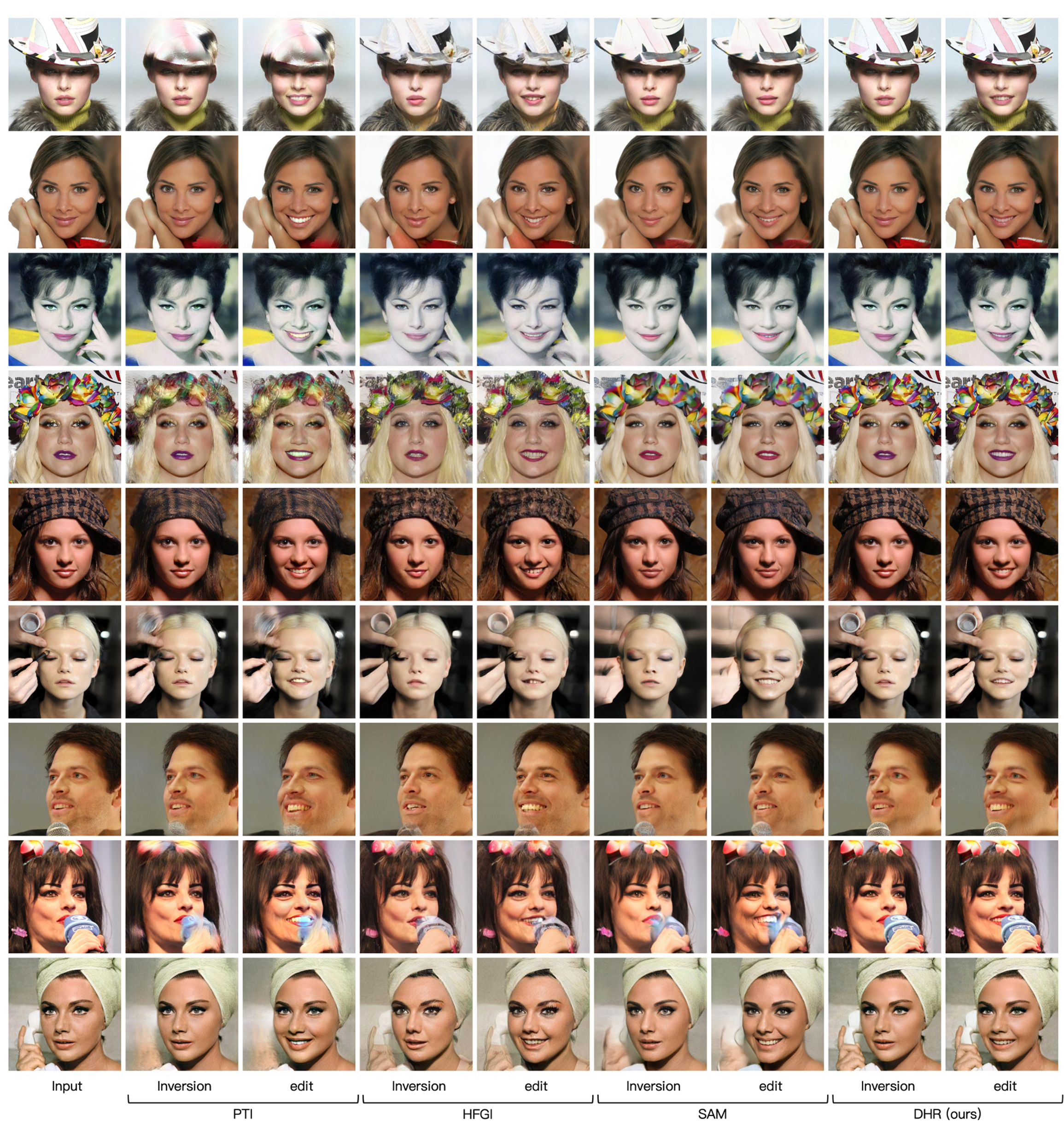}
	\caption{\textbf{Comparison of the editing result with the previous methods in ``Smile".}}
	\label{fig:compare_smile}
\end{figure*}

\begin{figure*}[t]
	\centering
	\includegraphics[width=1.0\linewidth]{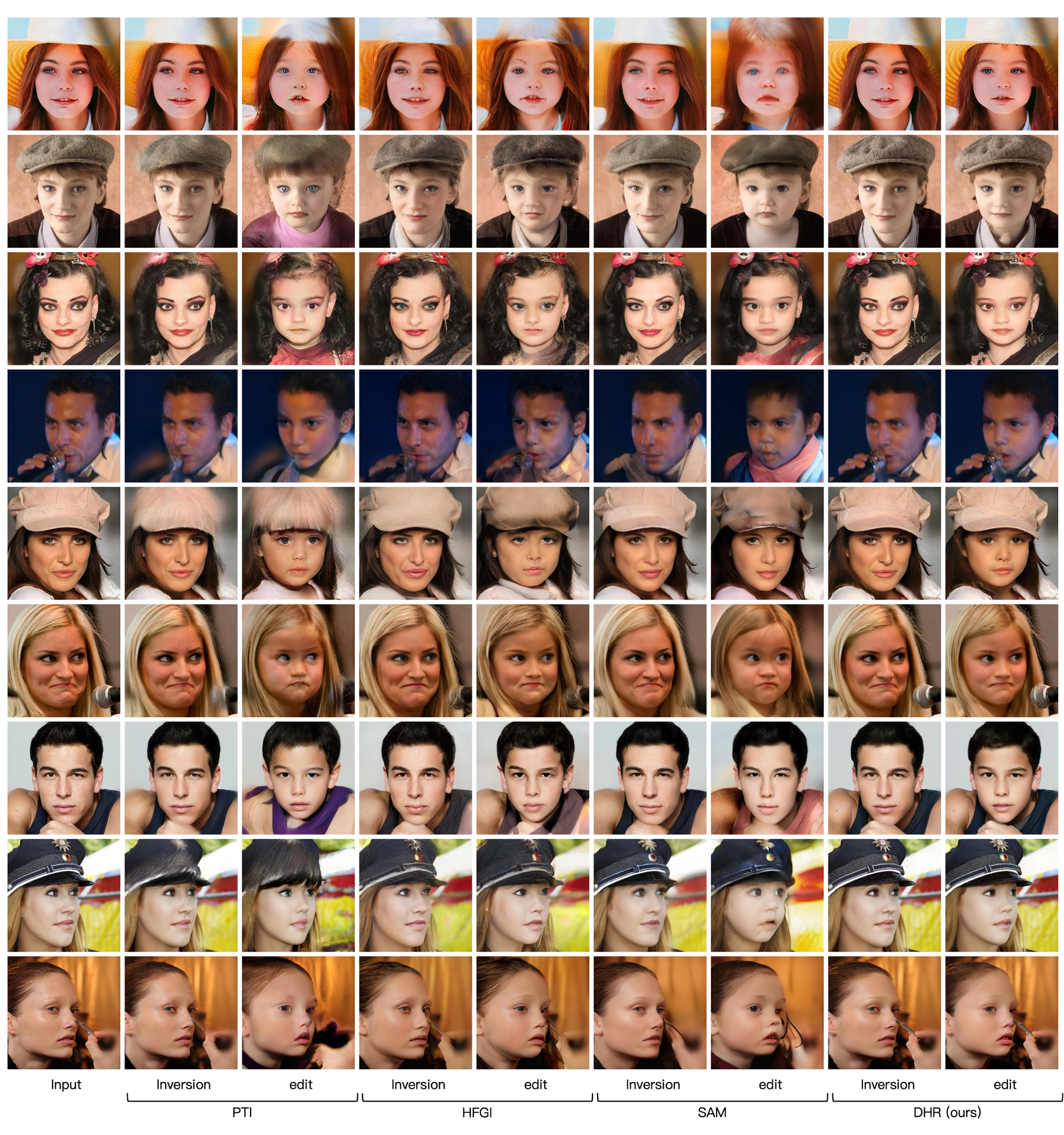}
	\caption{\textbf{Comparison of the editing result with the previous methods in ``Young".}}
	\label{fig:compare_young}
\end{figure*}

\begin{figure*}[t]
	\centering
	\includegraphics[width=1.0\linewidth]{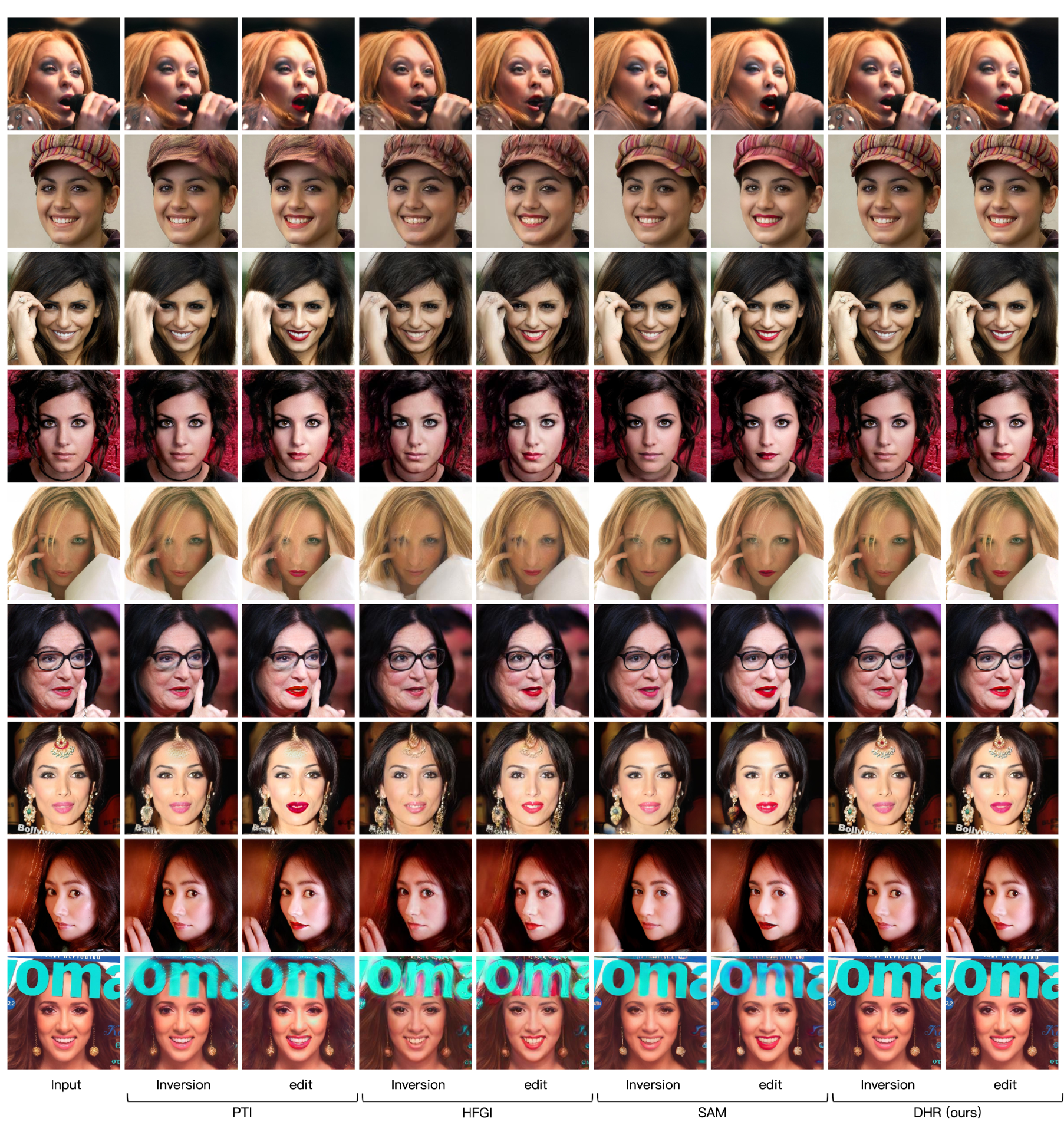}
	\caption{\textbf{Comparison of the editing result with the previous methods in ``Lipstick".}}
	\label{fig:compare_lipstick}
\end{figure*}

\begin{figure*}[t]
	\centering
	\includegraphics[width=1.0\linewidth]{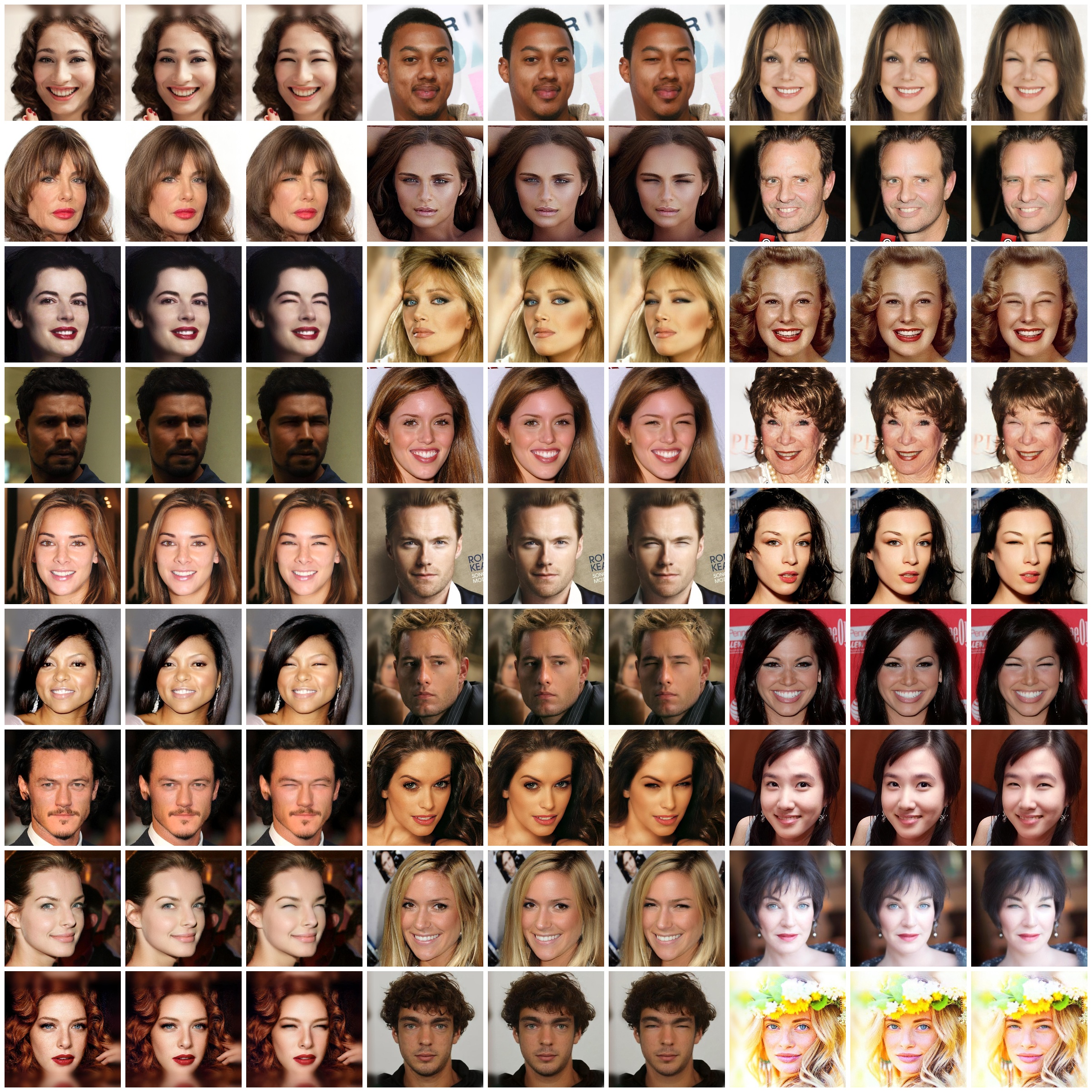}
	\caption{\textbf{Inversion and ``Eye Closure'' editing results.} The three columns represent \textbf{Input}, \textbf{Inversion}, and \textbf{Edit}.}
	\label{fig:eyeclose}
\end{figure*}

\begin{figure*}[t]
	\centering
	\includegraphics[width=1.0\linewidth]{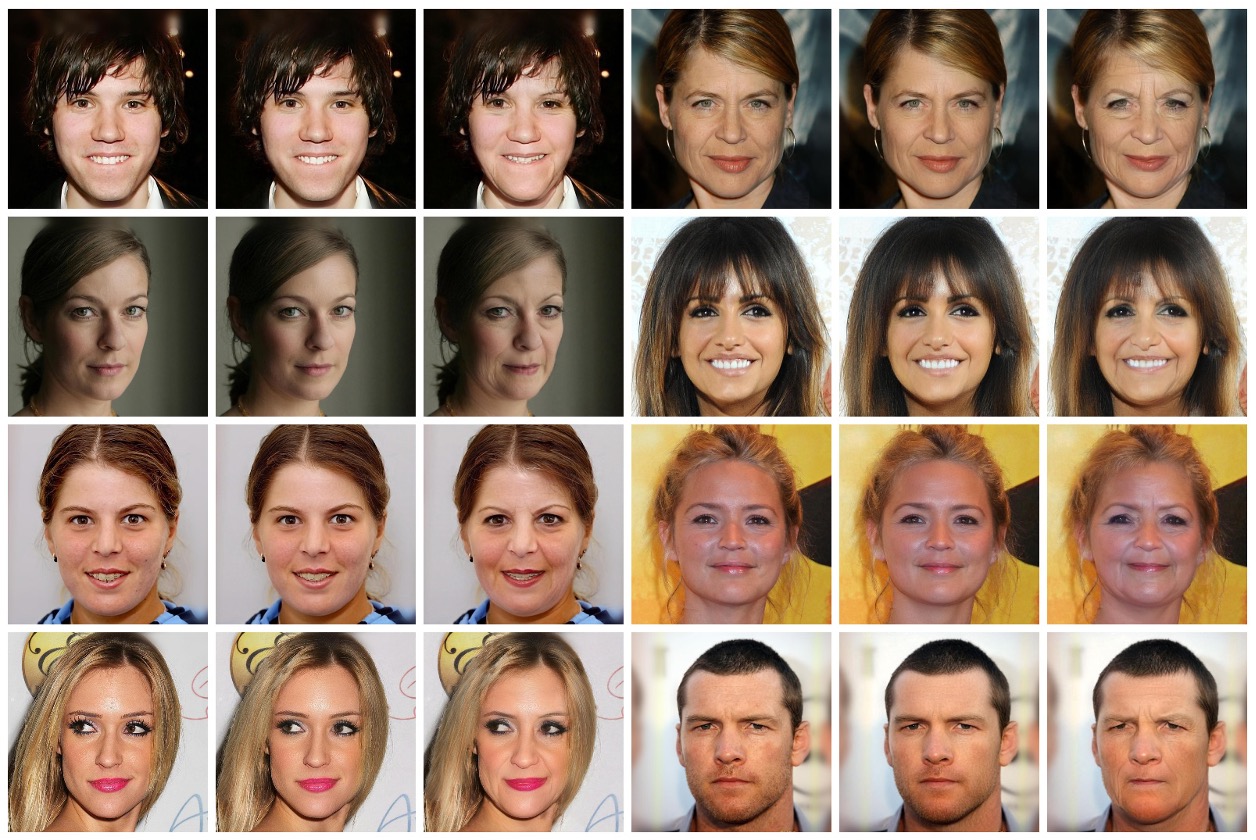}
	\caption{\textbf{Inversion and ``Wrinkles'' editing results.} The three columns represent \textbf{Input}, \textbf{Inversion}, and \textbf{Edit}.}
	\label{fig:wrinkles}
\end{figure*}

\begin{figure*}[t]
	\centering
	\includegraphics[width=0.9\linewidth]{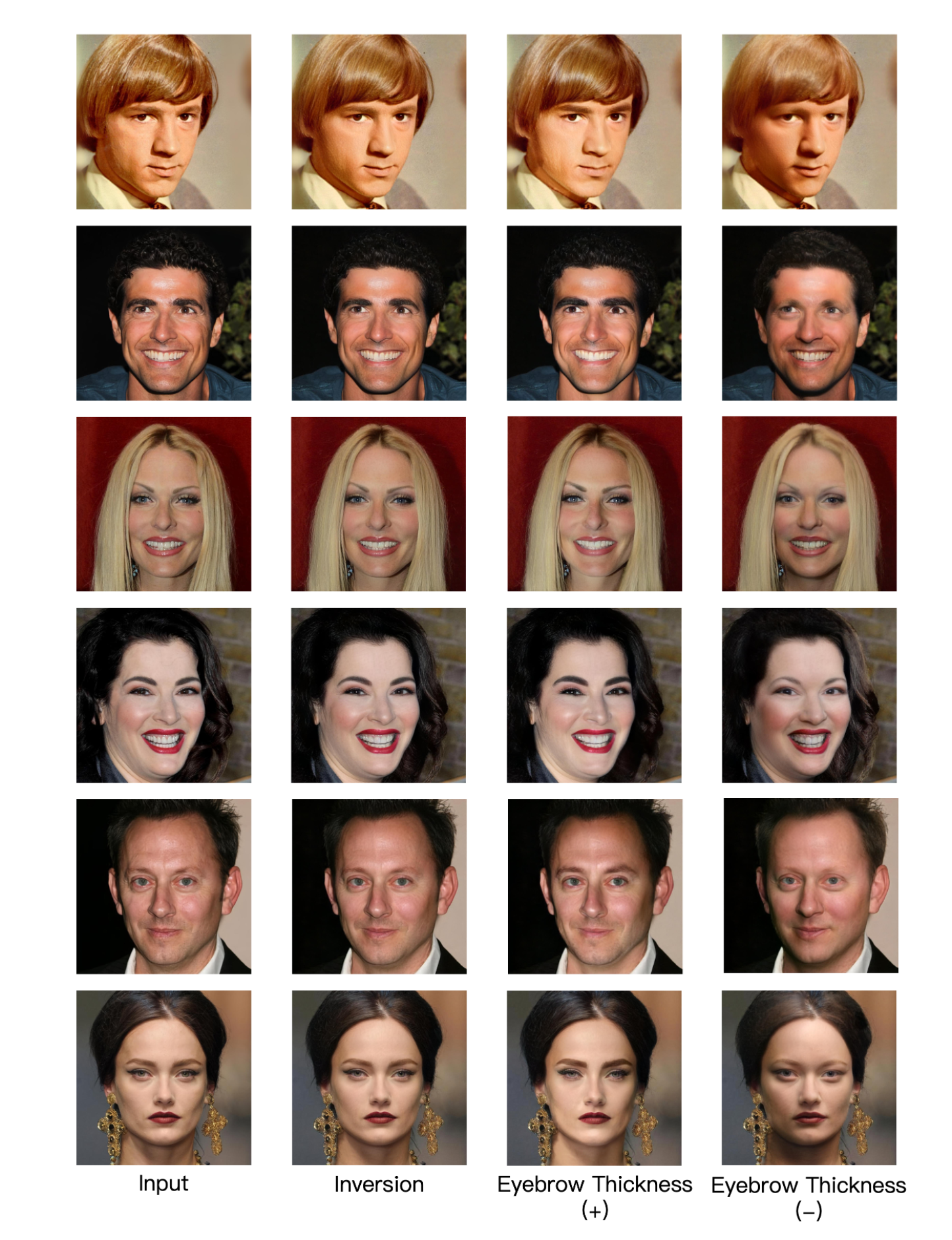}
	\caption{\textbf{Inversion and ``Eyebrow Thickness'' editing results.}}
	\label{fig:eyebrow}
\end{figure*}

We further demonstrate additional inversion and editing results:
\begin{itemize}
\item Figure~\ref{fig:compare_smile}: Comparison of ``Smile" Editing.
\item Figure~\ref{fig:compare_young}: Comparison of ``Young" Editing.
\item Figure~\ref{fig:compare_lipstick}: Comparison of ``Lipstick" Editing.
\item Figure~\ref{fig:eyeclose}: Results of ``Eye Closure" Editing.
\item Figure~\ref{fig:wrinkles}: Results of ``Wrinkles" Editing.
\item Figure~\ref{fig:eyebrow}: Results of ``Eyebrow Thickness" Editing.

\end{itemize}

\end{document}